\documentclass[10pt,conference,a4paper]{IEEEtran}
\IEEEoverridecommandlockouts
\usepackage{cite}
\usepackage{amsmath,amssymb,amsfonts}
\usepackage{algorithmic}
\usepackage{textcomp}
\usepackage{xcolor}
\def\BibTeX{{\rm B\kern-.05em{\sc i\kern-.025em b}\kern-.08em
    T\kern-.1667em\lower.7ex\hbox{E}\kern-.125emX}}

\usepackage[pdftex]{graphicx,epsfig}
\usepackage{url}
\usepackage[latin9]{inputenc}
\usepackage{mathrsfs,bm}
\usepackage{xspace}
\usepackage{color, colortbl}
\usepackage{soul}
\usepackage[caption=false]{subfig}
\usepackage{booktabs}
\usepackage{stfloats}
\usepackage{glossaries}
\usepackage{balance}
\usepackage{siunitx}
\usepackage{balance}
\usepackage{multirow}

\def\I{I}

\DeclareMathOperator*{\argmin}{arg\,min}

\newacronym{gan}{GAN}{Generative Adversarial Network}
\newacronym{dct}{DCT}{Discrete Cosine Transform}
\newacronym{mse}{MSE}{Mean Squared Error}
\newacronym{roc}{ROC}{Receiver Operating Characteristic}
\newacronym{cnn}{CNN}{Convolutional Neural Network}
\newacronym{auc}{AUC}{Area Under the Curve}
\newacronym{gpu}{GPU}{Graphics Processing Unit}
\newacronym{cgi}{CGI}{Computer-Generated Imagery}
\newacronym{qf}{QF}{Quality Factor}
\newacronym{fd}{FD}{First Digit}
\newacronym{pmf}{pmf}{Probability Mass Function}
\newacronym{fir}{FIR}{Finite Impulse Response}
\newacronym{iir}{IIR}{Infinite Impulse Response}
\newacronym{svm}{SVM}{Support Vector Machine}

\begin{document}

\title{On the use of Benford's law to detect GAN-generated images}

\author{\IEEEauthorblockN{Nicol\`o Bonettini}
\IEEEauthorblockA{\textit{DEIB} \\
\textit{Politecnico di Milano}\\
Milano, Italy \\
nicolo.bonettini@polimi.it}
\and
\IEEEauthorblockN{Paolo Bestagini}
\IEEEauthorblockA{\textit{DEIB} \\
\textit{Politecnico di Milano}\\
Milano, Italy \\
paolo.bestagini@polimi.it}
\and
\IEEEauthorblockN{Simone Milani}
\IEEEauthorblockA{\textit{Department of Information Engineering} \\
\textit{University of Padova}\\
Padova, Italy \\
simone.milani@dei.unipd.it}
\and
\IEEEauthorblockN{Stefano Tubaro}
\IEEEauthorblockA{\textit{DEIB} \\
\textit{Politecnico di Milano}\\
Milano, Italy \\
stefano.tubaro@polimi.it}
}

\maketitle

\begin{abstract}
The advent of \gls{gan} architectures has given anyone the ability of generating incredibly realistic synthetic imagery.
The malicious diffusion of \gls{gan}-generated images may lead to serious social and political consequences (e.g., fake news spreading, opinion formation, etc.).
It is therefore important to regulate the widespread distribution of synthetic imagery by developing solutions able to detect them.
In this paper, we study the possibility of using Benford's law to discriminate \gls{gan}-generated images from natural photographs.
Benford's law describes the distribution of the most significant digit for quantized \gls{dct} coefficients.
Extending and generalizing this property, we show that it is possible to extract a compact feature vector from an image.
This feature vector can be fed to an extremely simple classifier for GAN-generated image detection purpose.
\end{abstract}

\begin{IEEEkeywords}
image forensics, GAN, Benford's law
\end{IEEEkeywords}

\section{Introduction}\label{sec:intro}
With the advent of modern deep learning solutions such as \gls{gan}s, a new series of image and video editing tools has been made available to everyone (e.g., Recycle-GAN \cite{Bansal2018}, StyleGAN \cite{Karras2019}, etc.).
These techniques allow to synthesize realistic and visually-pleasant artificial images not resorting to complex \gls{cgi} techniques required in the past.
Unfortunately, this great step forward in technology came at a price.
Indeed, \gls{gan}s can be maliciously used by everyone to generate very realistic image forgeries to manipulate people's opinion through fake news spreading \cite{Brundage2018}.
To counter this threat, the forensic research community has started to develop a series of techniques that detect fake \gls{gan}-generated image \cite{Marra2018, Marra2019, mccloskey2019detecting}.

All of the above-mentioned strategies are among the latest solutions for \gls{gan} image detection.
However, the \gls{cgi} detection problem has been extensively investigated in the past multimedia forensic literature \cite{farid2012perceptual, Dang-Nguyen2015a, Rahmouni2017}.
It is worth noting that previous methods aimed at exposing some specific \gls{cgi} inconsistencies and artifacts from some characteristic statistical traces or according to a pre-defined model.
These strategies were suggested by the knowledge of the available \gls{cgi} algorithms that could have been applied to generate the fake image.
However, \gls{gan}-generated images can not be related to a well defined model, since each scheme presents its own peculiarities depending on the implemented architecture and training process. 
Indeed, as shown in \cite{Marra2019}, each architecture may introduce different traces, thus making their generalization a complex task.
A detector that has been trained to detect images generated by a specific \gls{gan} architecture could not be suitable for a different \gls{gan} scheme.

For this reason, the approach analyzed in this paper focuses on identifying and analyzing statistical traces that make \gls{gan}-generated images differ from natural photographs.
Previous work has shown that, on natural digital images,  the probability distribution of specific variables usually follows a pre-defined behavior that proves to be completely-altered whenever the image is modified.
As an example, the distribution of the first significant digit of quantized \gls{dct} coefficients follow Benford's law \cite{Perez-Gonzalez2007}.
This property can be proved whenever the statistics of quantized \gls{dct} coefficients shows an exponential decay and can be empirically-verified on real images. 

Exploiting this property, many forensics detectors have been successfully proposed in the literature (e.g., for detection of JPEG compression \cite{pasquini2014multiple, milani2014discriminating}, face morphing \cite{Makrushin2018}, \cite{acebo2015}  synthetic imagery, etc.).
Despite these premises, there is no current proof that \gls{gan}-generated pictures should be statistically compliant to Benford's law \cite{benford1938}.

In this work, we investigates whether Benford's law can be used for the detection of \gls{gan}-generated images.
The reported analysis is exploited to design a \gls{gan} image detector which proved to be extremely accurate with a limited computational effort.
More precisely, we verify that Benford's law is not followed by \gls{gan} images, and we propose a set of related features that could highlight this unfitting for an analyzed digital image.
A simple supervised learning framework is then proposed to detect if an image is natural or \gls{gan}-generated from the extracted features.

This solution is evaluated on an image corpus made available by the authors of \cite{marra2019incremental}, enriched by additional images obtained by more modern \gls{gan}s.
We make use of more than \num{200000} \gls{gan}-generated images obtained through different architectures trained on different tasks on different datasets.
Results show that there is a trade-off between the chosen size of the proposed feature vector and the achieved accuracy.
It is possible to either use a compact feature vector to obtain results comparable with the state-of-the-art, or a larger feature vector that allows improving against the more recent solutions proposed in the literature.
This flexibility makes the proposed solution particularly suitable also for low-power devices not equipped with an advanced \gls{gpu}, which might still need to detect whether images are fake or not (e.g., smartphones, tablets, etc.).
Additionally, we discuss resilience to JPEG compression in order to better define the working conditions of the proposed method.


\section{Background}\label{sec:background}
\begin{figure}[t]
	\centering
	\includegraphics[width=.6\columnwidth]{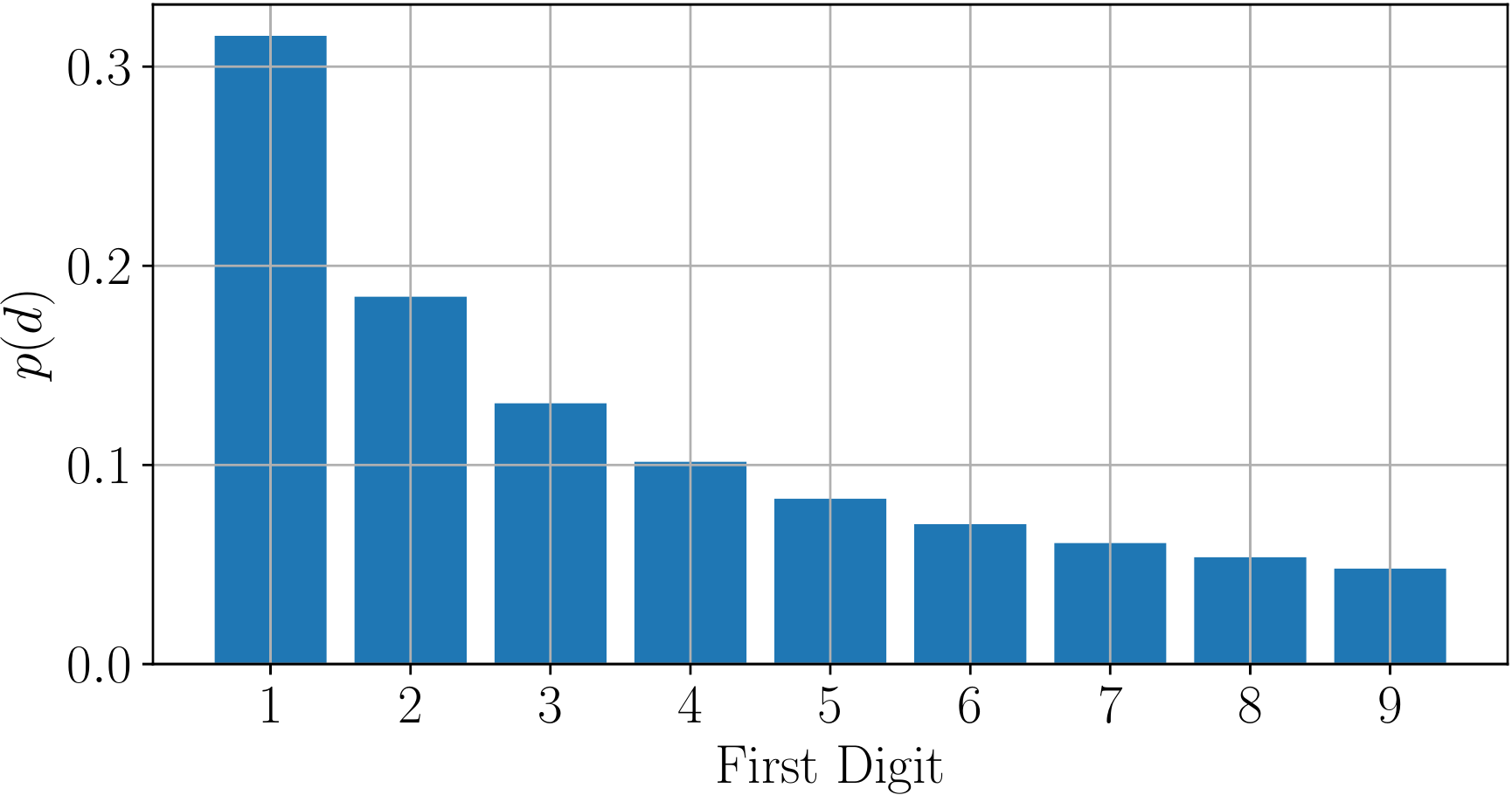}
	\caption{Benford's law \gls{fd} \gls{pmf} considering base \num{10} for \gls{fd} computation.}
	\label{fig:benford}
\end{figure}

Benford's law, which is also known as First Digit (\gls{fd}) law or Significant Digit law, concerns the statistical frequencies  of the most significant digits for the elements of a real-life numerical set.
More precisely, the rule states that, given a set of measurements for some natural quantities (e.g., population of cities, stock prices, etc.), the statistics of their \gls{fd} follows the distribution depicted in Fig.~\ref{fig:benford} and described by the equation
\begin{equation}
	p(d) = \log_{10} \left( 1 + \frac{1}{d}\right),
\end{equation}
where $d$ is the \gls{fd} in base \num{10} (the generalized version of this law is presented in the next section).
This has been empirically-observed  over a vast range of natural quantities~\cite{raimi1976}, but it is also possible to prove it in closed form for many exponentially-decreasing probability distributions \cite{Wan09:benford}.
It has also been observed that this rule is not well-fitted by \gls{fd} statistics from altered data: whenever numbers are changed according to some selective strategies, \gls{fd} frequencies deviate from their theoretical values~\cite{diekmann2007}.
As a consequence, this proof has been used as supporting evidence for detecting falsified accounts, fake financial reports, and frauds \cite{Tod09:benford}.

This property has been largely exploited in multimedia forensics to detect image tampering.
In fact, natural image \gls{dct} coefficients can be typically modeled by a Laplacian-like distribution~\cite{smoot1998}, which naturally follow Benford's law, and for this reason, the mentioned rule can be successfully used in image forensic applications~\cite{perez2015book}.

A well-known application of Benford's law in forensics is the study of JPEG compression traces \cite{Pev08:benford}: the authors propose using such rule to verify if an image has been JPEG compressed once or twice.
Milani et al.~\cite{milani2014discriminating} exploit \gls{fd}'s features to detect multiple JPEG compression, also showing robustness against rotation and scaling.
Pasquini et al.~\cite{pasquini2014multiple} address the multiple JPEG compression detection problem by means of Benford-Fourier analysis.
The same authors also investigate traces of previous hidden JPEG compression in uncompressed images \cite{Pasquini2017}.

This rule has also been successfully applied to other forensic problems.
In \cite{milani2016phylogenetic}, the authors show that it is possible to leverage \gls{fd} distribution to roughly estimate the amount of processing that has been applied to a given image. 
The authors of \cite{moin2017benfords} apply Benford's law to solve image contrast enhancement detection problem.
In \cite{matern2019exploiting}, the authors make use of this law to deal with splicing forgery localization.

Another interesting application of Benford's law in image forensics is detecting computer graphics and computer generated images.
To this purpose, Del Acebo et al.~\cite{acebo2015} model light intensity in natural and synthetic images, concluding that FD's law is not followed by the latter.
Makrushin et al.~\cite{Makrushin2018} show how to efficiently detect morphed faces using the fitting parameters of the Benford's logarithmic curve as a features.

Anyway, detecting synthetic images is nowadays a timely and crucial forensic need due to the achievements of \gls{gan} technology in generating highly-realistic fake photographs.
This possibility has been recently used to create false image and video contents in deepfake political propaganda, revenge porn, fake news creation.
For these reasons, during the last years multimedia forensics researchers have been focusing on designing reliable strategies to detect synthetic images.

To this purpose, \cite{Marra2018} proposes a method to detect image-to-image translation over social networks.
Specifically, the authors compare different detectors fine-tuned for the binary classification task of \gls{gan}-generated against natural image detection.
The same authors also show how a model-specific fingerprint can be retrieved by \gls{gan} generated images in order to identify the specific network used for image generation \cite{Marra2019}.
In \cite{marra2019incremental}, authors apply an incremental learning strategy to train a \gls{gan}-generated image detector that can be progressively updated in time as new images from different kinds of  \gls{gan}s are processed.
In \cite{li2018disparity}, the authors propose a method to detect \gls{gan}-generated images by analyzing the disparities in color components between real scene images and generated images.
In \cite{mccloskey2019detecting}, \gls{gan} images are detected by analyzing saturation artifacts in pixel distributions.
Moreover, if videos are analyzed, methods exploiting also the temporal evolution of frames have been proposed \cite{Guera2019, matern2019exploiting}.

\section{Motivations}\label{sec:motivations}
Natural images, as many other natural processes, can be roughly approximated as autoregressive signals \cite{delp1979image}.
This is the rationale behind different historical as well as more recently proposed image compression \cite{delp1979image, minnen2018joint} and generation \cite{oord2016pixel, oord2016conditional} methods.
From these assumptions, an image can be modeled as a complex autoregressive signal with a generally low-pass characteristics.

\gls{gan} generator's structures are usually composed by a concatenation of limited-support convolutional layers followed by non-linearities.
Filters' coefficients are optimized so that \gls{gan}'s response to a given input belongs to the desired output class.
However, in most \gls{gan} implementations,  practical and complexity reasons have led to the adoption of filters with a limited size.
Therefore, if no recursive operations are applied in the network architecture, the output of a \gls{gan} generator looks more like a signal filtered through a \gls{fir} filter than a complex autoregressive process.


The rationale behind the proposed method is that the information related to the filter ideally used to generate the data under analysis can be used to discriminate natural images (with autoregressive and complex spectra) from \gls{gan}-generated ones (generated through operations closer to \gls{fir} filtering).
This can be done analyzing the statistics of quantized \gls{dct} coefficients.

More precisely, let us assume that an input grayscale image is partitioned into $K$ distinct $8 \times 8$ blocks, which are then mapped into the 2D-\gls{dct} domain and further quantized.
This processing chain is used by the JPEG coding standards and proves to be tailored to the spectral characteristics of images.
Some of the past works highlight that, in the frequency domain, the quantized \gls{dct} coefficient statistics of natural images must follow Benford's law \cite{perez2015book}.

Let us denote as $c_{n, \Delta}(k)$ the DCT coefficient at the $n$-th frequency in zig-zag mode obtained from the $k$-th block and quantized with step $\Delta$.
It is possible to compute the corresponding \gls{fd} with base $b$ as
\begin{equation}
	d_{b, n, \Delta}(k) = \left\lfloor \frac{|c_{n,  \Delta}(k)|}{b^{\left\lfloor \log_b |c_{n,  \Delta}(k)|  \right\rfloor}}   \right\rfloor.
	\label{eq:fd}
\end{equation}

As $d_{b, n, \Delta}(k)$ can only assume $b-1$ values (i.e., all possible digits defined in base $b$ apart from zero), its \gls{pmf} $\hat{p}_{b, n, \Delta}$ computed over the $K$ blocks is composed by $b-1$ elements.
For the sake of notational compactness, let us momentarily drop the indexes $n$, $b$, and $\Delta$.
We can formally define the \gls{pmf} $\hat{p}(d)$ as
\begin{equation}
	\hat{p}(d) = \frac{1}{K} \sum_{k=1}^K \mathbf{1}_x(d(k)), \; d \in \{1, 2, \ldots, b-1\},
	\label{eq:p_hat}
\end{equation}
where
\begin{equation}
	\mathbf{1}_x(y) =
	\begin{cases}
	1 & \text{if} \, y = x,\\
	0 & \text{otherwise}.
	\end{cases}
\end{equation}
This \gls{pmf} for a natural image must follow the generalized Benford's law equation
\begin{equation}
	p(d) =  \beta \log_{b} \left( 1 + \frac{1}{\gamma + d^\delta} \right),
	\label{eq:p}
\end{equation}
where $\beta$ is a scale factor, $\gamma$ and $\delta$ parameterize the logarithmic curve, and $d \in \{1, 2, ..., b-1\}$ is one possible value of the considered first digits in base $b$.


The fitness between $\hat{p}(d)$ and $p(d)$ can be measured by some divergence functions such as the Jensen-Shannon divergence $D^\text{JS}\left(\hat{p}  | p \right)$
\begin{equation}
	D^\text{JS}\left(\hat{p}  | p \right) = D^\text{KL}\left( \hat{p} | p  \right) + D^\text{KL}\left(p | \hat{p} \right),
\end{equation}
which is a symmetrized version of the well-known Kullback-Leibler divergence
\begin{equation}
	D^\text{KL}\left( \hat{p} | p  \right) =  \sum_{d=1}^{b-1}  \hat{p}(d) \log \frac{ \hat{p}(d)}{p(d)}.
\end{equation}

Since $D^\text{JS}$ proves to be unstable for biased \gls{pmf}s, it is possible to use the symmetrized Renyi divergence
\begin{equation}
	D^\text{R}_{\alpha}\left(\hat{p}  | p \right)  = \displaystyle \frac{1}{1-\alpha}  \left( \log  S_\alpha\left(\hat{p} ,p \right)  + \log    S_\alpha\left( p, \hat{p} \right) \right),
\end{equation}
or the symmetrized Tsallis divergence
\begin{equation}
	D^\text{T}_{\alpha}\left(\hat{p}  | p \right) = \displaystyle \frac{1}{1-\alpha}  \left(  2 - S_\alpha\left(\hat{p} ,p \right)   - S_\alpha\left(p,\hat{p}\right) \right),
\end{equation}
where
\begin{equation}
	S_\alpha\left(q, p\right)  = \sum_{d=1}^{b-1} q(d)^\alpha / p(d)^{\alpha-1}.
\end{equation}

%

It is possible to prove that, whenever an image is altered (e.g., it is compressed/quantized a second time, etc.), Benford's law is not verified anymore.
In fact, many modifications redistribute image coefficients among the bins of the quantizer, thus the final \gls{pmf} associated to quantized \gls{dct} coefficients presents some oscillating probability values that deviate from the ideal distribution.
For these reasons, many solutions in the past measured the divergence between the empirically-estimated $\hat{p}(d)$ and its ideal fitted version $p(d)$ in order to find whether the image has been altered or not.
In this paper we show that it is possible to adopt the same solution to detect \gls{gan}-generated pictures.

\section{GAN-generated image detection}\label{sec:method}
In this section we provide a formal definition of the \gls{gan}-generated image detection problem and report all the technical details about the detection method we propose.

\subsection{Problem formulation}
We define the \gls{gan}-generated image detection problem as a two-class classification problem.
Given an image $\I$, we want to understand whether it has been synthetically generated through a \gls{gan}, or it is a natural photograph.

Formally, to solve this problem we consider a pipeline composed by two blocks: a feature extractor and a supervised classifier.
The feature extractor implements the function $\Phi(\cdot)$, which turns the image into a more compact yet informative representation, i.e., the feature vector $\bm\phi = \Phi(\I)$.
The classifier implements the function $M(\cdot)$ such that: $M(\bm\phi) = 0$ if the image is a natural one; $M(\bm\phi) = 1$ if the image comes from a \gls{gan}.
With this framework in mind, we focus on designing the function $\Phi(\cdot)$ based on Benford's law, so that a simple classifier can be effectively used.

\subsection{Detection method}
\begin{figure}
	\centering
	\includegraphics[width=\columnwidth]{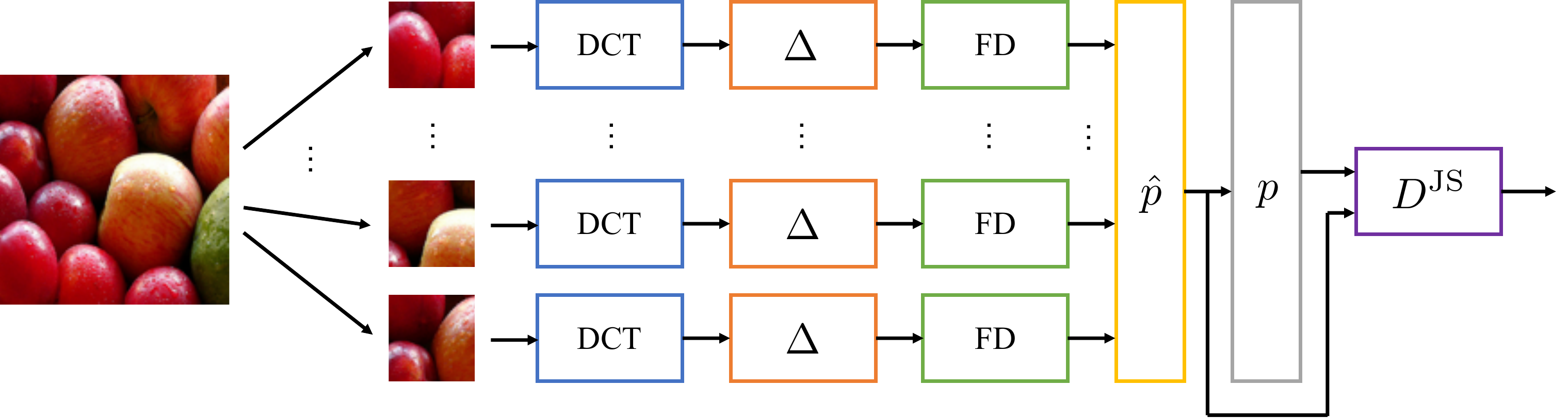}
	\caption{Feature extraction pipeline considering a single divergence, quantization step $\Delta$, base $b$ and \gls{dct} coefficient $n$. Extraction process is repeated for multiple parameters.}
	\label{fig:pipeline}
\end{figure}

The feature extraction process is depicted in Fig.~\ref{fig:pipeline}.
Given an image $\I$, we divide it in $K$ non-overlapping blocks with resolution $8 \times 8$ pixel.
From each block, we compute its 2D-\gls{dct} representation.
We then quantize it using a given quantization step $\Delta$ (chosen for each coefficient according to a JPEG quantization matrix).

\begin{figure}
	\centering
	\subfloat[$b=10$]{\includegraphics[width = \columnwidth]{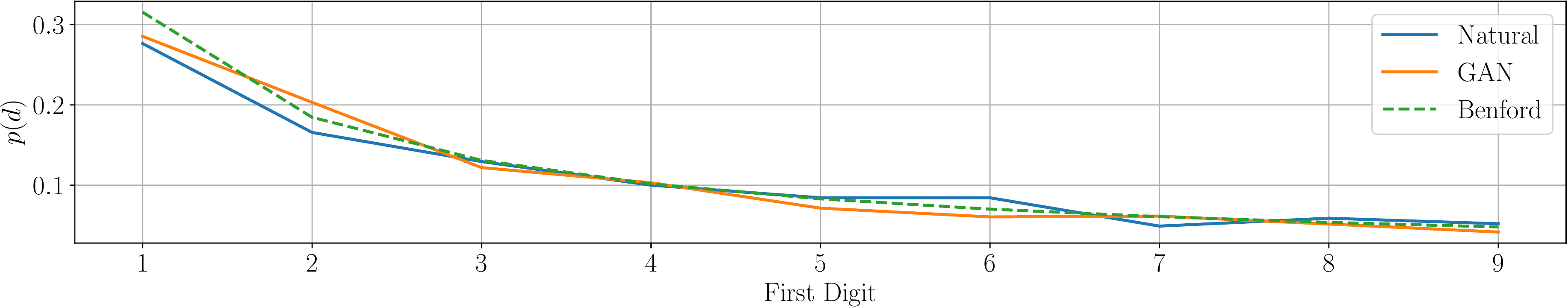}} \\
	\subfloat[$b=40$]{\includegraphics[width = \columnwidth]{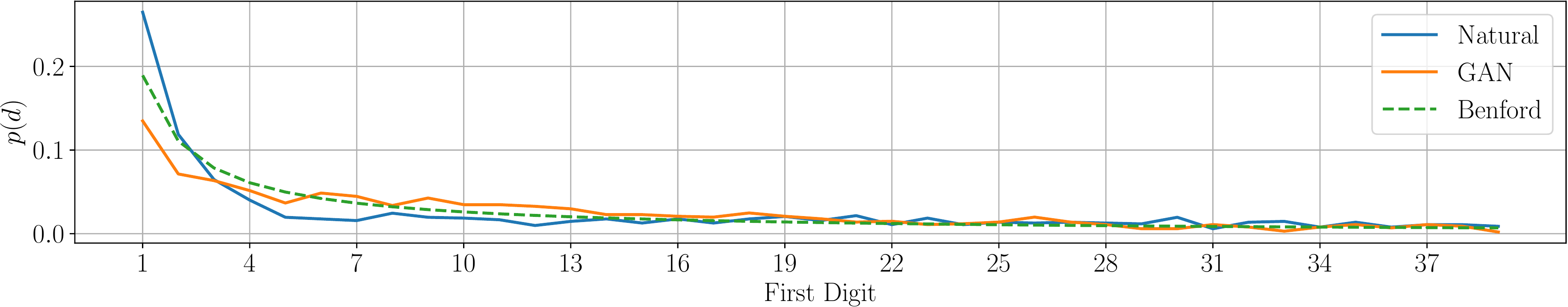}} \\
	\caption{Different \gls{pmf} $\hat{p}$ for natural (blue) and \gls{gan}-generated images (oranges) are compared to the ideal Benford curve (dashed green) for different bases $b$. Blue and orange curves deviates differently from the green one.}
	\label{fig:histograms}
\end{figure}

Given a base $b$, we compute the first digit of the $n$-th quantized 2D-\gls{dct} frequency sample from the $k$-th block according to \eqref{eq:fd}.
We then compute the \gls{pmf} $\hat{p}_{b, n, \Delta}$ according to \eqref{eq:p_hat}.
Examples of $\hat{p}$ for different bases for both natural and \gls{gan}-generated images are reported in Fig.~\ref{fig:histograms}.
Finally, we fit generalized Benford's law expressed in \eqref{eq:p} by solving a mean square error minimization problem as
\begin{equation}
	p^\text{fit}_{b,n,\Delta} = \argmin_p \sum_{d=1}^{b - 1}(\hat{p}_{b,n,\Delta}(d) - p(d))^2.
\end{equation}

Comparing the computed \gls{pmf} $\hat{p}_{b, n, \Delta}$ and the Benford fit $p^\text{fit}_{b,n,\Delta}$, we compute Jensen-Shannon divergence $D^\text{JS}_{b, n, \Delta}$, Renyi divergence $D^\text{R}_{b, n, \Delta}$, and Tsallis divergence $D^\text{T}_{b, n, \Delta}$ as reported in Section~\ref{sec:motivations}.
Notice that we removed the dependency of Tsallis and Renyi divergence on $\alpha$ as we keep it constant in our experiments.

Finally, considering a set $\mathcal{B}$ of bases, a set $\mathcal{N}$ of \gls{dct} frequencies and a set $\mathcal{J}$ of JPEG quality factors driving the quantization parameter $\Delta$, we obtain the final feature vector by concatenating all divergences as:
\begin{equation}
	\bm{\phi}_{\mathcal{B}, \mathcal{N}, \mathcal{J}} = [D^\text{JS}_{b, n, \Delta}, \, D^\text{R}_{b, n, \Delta}\, D^\text{T}_{b, n, \Delta}]_{b \in \mathcal{B}, n \in \mathcal{N}, \Delta \in \mathcal{J}}.
\end{equation}
Notice the the feature vector size depends on how many \gls{dct} coefficients, bases and quantization steps are used during the analysis.
For instance, if we choose a single compression step, a single \gls{dct} frequency and a single base, the feature vector will be composed by the concatenation of just three divergences, thus having dimensionality \num{3}.
Conversely, if we use multiple bases, frequencies and compression steps, we end up with a bigger vector.
In our experiments, we consider vectors with dimensionality ranging from \num{3} to \num{540}, as shall be explained in Section~\ref{sec:results}.

After feature computation, the vector $\bm{\phi}_{\mathcal{B}, \mathcal{N}, \mathcal{J}}$ is fed to a supervised classifier.
In order to study the effectiveness of Benford-based features, we do not adopt unnecessarily complicated classifiers $M(\cdot)$.
Specifically, we resort to a Random Forest classifier.

\section{Results}\label{sec:results}
In this section we discuss the used datasets, the experimental setup, and finally report all the results achieved with the proposed technique for \gls{gan}-generated image detection.

\subsection{Dataset}
\begin{figure}[t]
	\centering
	\subfloat{\includegraphics[width=0.15\linewidth]{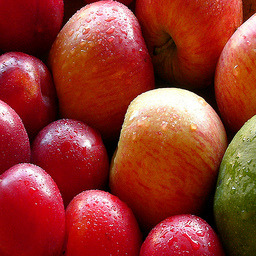}} \hfil
	\subfloat{\includegraphics[width=0.15\linewidth]{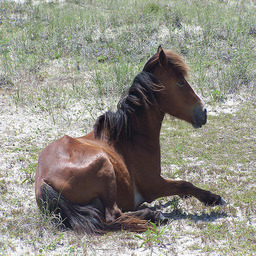}} \hfil
	\subfloat{\includegraphics[width=0.15\linewidth]{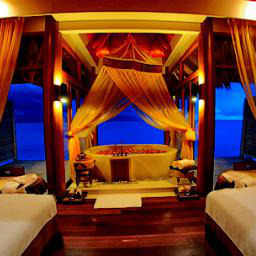}} \hfil
	\subfloat{\includegraphics[width=0.15\linewidth]{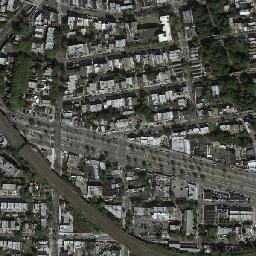}} \hfil
	\subfloat{\includegraphics[width=0.15\linewidth]{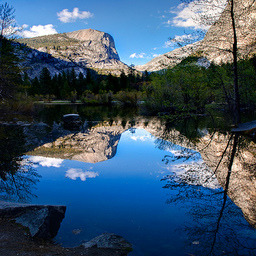}} \hfil
	\subfloat{\includegraphics[width=0.15\linewidth]{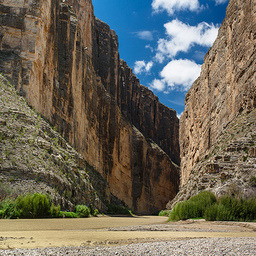}} \\
	\vspace{-.8em}
	\subfloat{\includegraphics[width=0.15\linewidth]{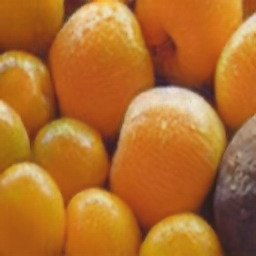}} \hfil
	\subfloat{\includegraphics[width=0.15\linewidth]{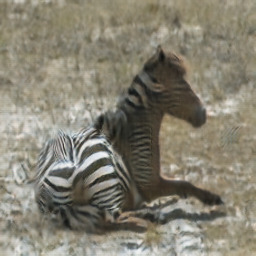}} \hfil
	\subfloat{\includegraphics[width=0.15\linewidth]{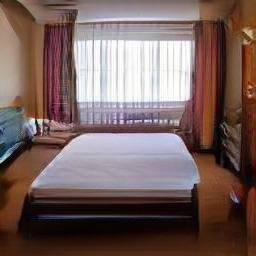}} \hfil
	\subfloat{\includegraphics[width=0.15\linewidth]{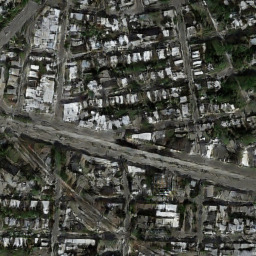}} \hfil
	\subfloat{\includegraphics[width=0.15\linewidth]{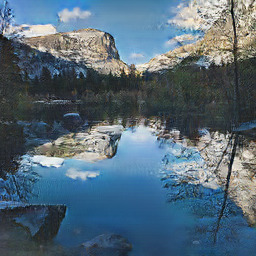}} \hfil
	\subfloat{\includegraphics[width=0.15\linewidth]{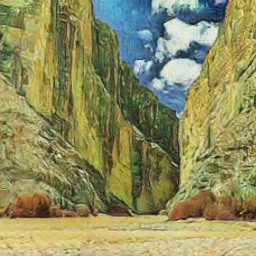}} \hfil
	\caption{Examples of original (top) and \gls{gan}-generated (bottom) images belonging to the dataset proposed in \cite{marra2019incremental}.}
	\label{fig:example_images}
\end{figure}

In order to build our dataset, we started from the publicly available \gls{gan}-dataset released by Marra et al.~\cite{marra2019incremental}.
Specifically, we considered a corpus composed by \num{15} different sub-datasets of images obtained employing $2$ different architectures: Cycle-Gan~\cite{Zhu2017} and ProGAN~\cite{karras2018}.
The first architecture is designed for image-to-image translation purpose, i.e., mapping an image of a given class (i.e., pictures of horses) to an image of another one (i.e., pictures of zebras).
The second architecture is a generator able of creating natural looking pictures of different scenes depending on the used training data (e.g., bedroom pictures, bridges, etc.).
Each dataset comprises both natural images and their \gls{gan}-generated counterparts.
All images are color images and have a resolution of \num{256 x 256} pixel.
The complete composition is reported in Table~\ref{tab:dataset} and some examples of the more than \num{200000} images are reported in Fig.~\ref{fig:example_images}.

\begin{table}[t]
	\centering
	\caption{Dataset composition}
	\label{tab:dataset}
	\begin{tabular}{ccc}
		\toprule
		\textbf{Architecture} & \textbf{Dataset}    & \textbf{Number of images} \\ \midrule
		                      & orange2apple        & $1280 $                     \\
		                      & photo2ukiyoe        & $4072 $                     \\
		                      & winter2summer       & $1484 $                     \\
		                      & zebra2horse         & $1670 $                     \\
		Cycle-Gan             & photo2cezanne       & $3978 $                     \\
		                      & photo2vangogh       & $4099 $                     \\
		                      & photo2monet         & $4765 $                     \\
		                      & facades             & $259  $                     \\
		                      & cityscapes          & $1996 $                     \\
		                      & sats                & $684  $                     \\ \midrule
		                      & lsun\_bedroom       & $30770$                     \\
		                      & lsun\_bridge        & $28768$                     \\
		ProGAN                & lsun\_churchoutdoor & $29120$                     \\
		                      & lsun\_kitchen       & $42706$                     \\
		                      & lsun\_tower         & $29020$                     \\ \bottomrule
	\end{tabular}
\end{table}

\subsection{Setup}
As shown in Section~\ref{sec:method}, each feature depends on a selected set of bases $\mathcal{B}$, \gls{dct} frequencies $\mathcal{N}$, and analysis JPEG \gls{qf} describing the set of quantization steps $\mathcal{J}$.
With regards to bases, we test all combinations of sets of bases $\mathcal{B} \subseteq \{10, 20, 40, 60\}$ containing from one to four elements.
This leads to $15$ possible combinations of bases (i.e., from $\mathcal{B} = \{10\}$ to $\mathcal{B} = \{10, 20, 40, 60\}$).
Concerning the selected \gls{dct} frequencies, we choose a limited amount of sets $\mathcal{N} \subseteq \{1, 2, \ldots, 9\}$ (i.e., including the first 9 frequencies in zig-zag order after DC)
, similarly to other detectors available in literature~\cite{milani2014discriminating}).
Specifically, we only consider 9 sets obtained by progressively-adding one frequency at a time from the previous set (i.e., $\mathcal{N} = \{1\}$, $\mathcal{N} = \{1, 2\}$, $\ldots$, $\mathcal{N} =  \{1, 2, \ldots, 9\}$).
As for the quantization step values, the final feature set was created concatenating the \num{5} arrays of divergences obtained using JPEG \gls{qf}s in the set  $\mathcal{J} \subseteq \{80, 85, 90, 95, 100\}$.
Considering all combinations of bases, frequencies and JPEG quantization steps, we obtain a total amount of \num{675} different setups.

\begin{figure}[t]
	\centering
	\includegraphics[width = 0.7\columnwidth]{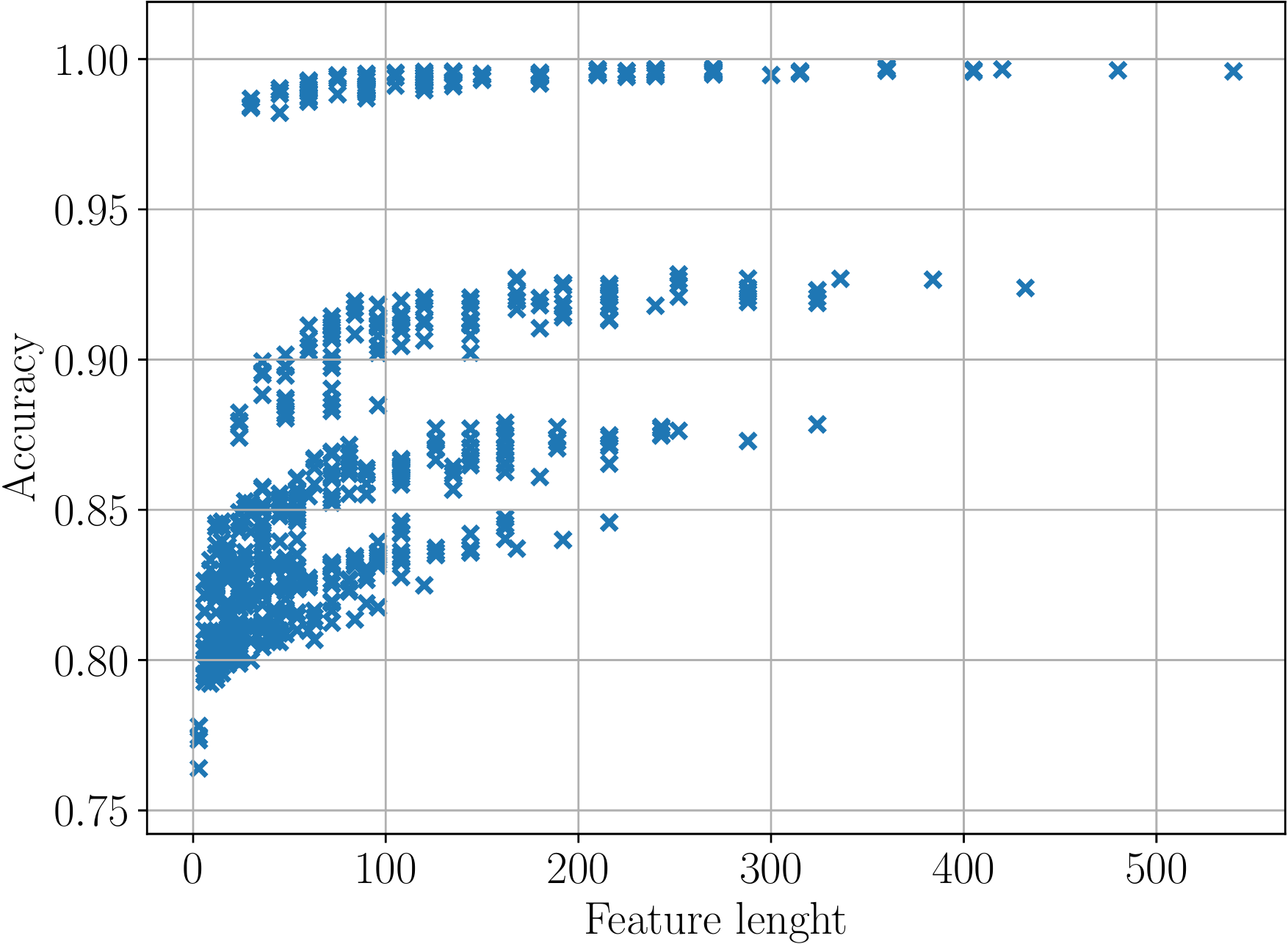}
	\caption{Accuracy obtained with different feature vectors obtained changing the considered sets $\mathcal{B}$,  $\mathcal{N}$ and $\mathcal{J}$. Each vector has a different length and provides a different accuracy result.}
	\label{fig:feat_len}
\end{figure}

For each feature vector described in Section~\ref{sec:method} (i.e., each setup), we trained a different Random Forest classifier performing Leave-One-Group-Out cross-validation over the various datasets as explained in \cite{Marra2018}.
The choice of Random Forest was suggested by the  low complexity requirements, the generalization capabilities, and the resilience to small training datasets. 
Namely, given a dataset $\mathcal{D}_i$ out of the complete set of dataset $\mathcal{D}$, we trained our model over the remaining $\mathcal{D}_j$, $\forall j \neq i$ and we test over $\mathcal{D}_i$.
Results are always shown on the leave-out dataset, and we report the maximum accuracy value among all the different setups.
To provide a practical example, let us consider the situation in which we test on dataset $\mathcal{D}_i = \texttt{orange2apple}$.
This consists of original images (i.e., apples and oranges) and \gls{gan} images (apples turned into oranges and viceversa).
The classifier was trained on all the other images (excluding those in \texttt{orange2apple}) in order to avoid biasing the results with overfitting.
We adopted the  Random Forest implementation provided with the open source Scikit Learn Python library. 
After a grid search over several candidates, we fixed the number of Decision Trees to $100$, with bootstrap sampling enabled.
We selected Gini index as splitting policy, leaving all the other parameters as their default values.


\subsection{Experiments}
\label{sec:experiments}
In this section we report all experimental results achieved to evaluate the proposed technique.
Moreover, we report a comparison against baseline solutions.
Finally, we provide some additional insights in terms of resilience to JPEG compression.

To select the baselines, we focused on the work proposed by Marra et al.~\cite{Marra2018} since, to the best of our knowledge, it is the only work to perform an extensive \gls{gan} detection test over a large dataset of images. Specifically, we selected two baselines: a completely data-driven one based on deep learning; a solution based on hand-crafted features commonly used in the forensics literature.

Similarly to the solution in~\cite{Marra2018}, we compared our approach with the Xception \gls{cnn}, as the first baseline method.
According to the results in~\cite{Marra2018}, this set of features seems to  provide the best results over most of the considered datasets.
Starting from the pretrained model, we finetuned it on our dataset, following the same Leave-One-Group-Out strategy we adopted for the Random Forest training. We used $70\%$ of the training data for the actual training, and the remaining $30\%$ for validation, testing on the Leave-Out dataset. We resorted to Adam optimization algorithm, with an initial learning rate of $0.0001$, training until reaching convergence on a validation plateau.
We adopted the Keras implementation of Xception, performing the training in several hours on a workstation equipped with a NVIDIA Titan V \gls{gpu}, a Intel Xeon E5-2687W and 256 GB of RAM.

The second baseline method  operates a linear Support Vector Machine (SVM) on a set of  handcrafted steganalysis rich-features (as suggested in~\cite{Marra2018}). These features have been successfully used in image forgery detection tasks as well~\cite{cozzolino2014}.
The model has been trained following the same train/test strategy used for Xception, using the Scikit Learn implementation of \gls{svm}.

\vspace{.5em}\noindent\textbf{Feature length and parameters.}
In the design of the proposed solution we considered different combinations of features obtained varying the parameters in the set $\mathcal{B}$, $\mathcal{N}$, $\mathcal{I}$ and changing feature vectors' lengths. 
As a matter of fact, it is necessary to evaluate how the vector length could impact on the classifier performance.
Fig.~\ref{fig:feat_len} shows the average test accuracy obtained on all datasets considering all possible \num{675} feature vectors.
It is possible to notice that even the smallest feature vectors of just \num{3} elements enable achieving an accuracy greater than $0.75$.
It is sufficient to use \num{50} features to have accuracy higher than $0.97$.


\begin{figure}[t]
	\centering
	\subfloat[single $\Delta$ value]{\includegraphics[width = 0.48\columnwidth]{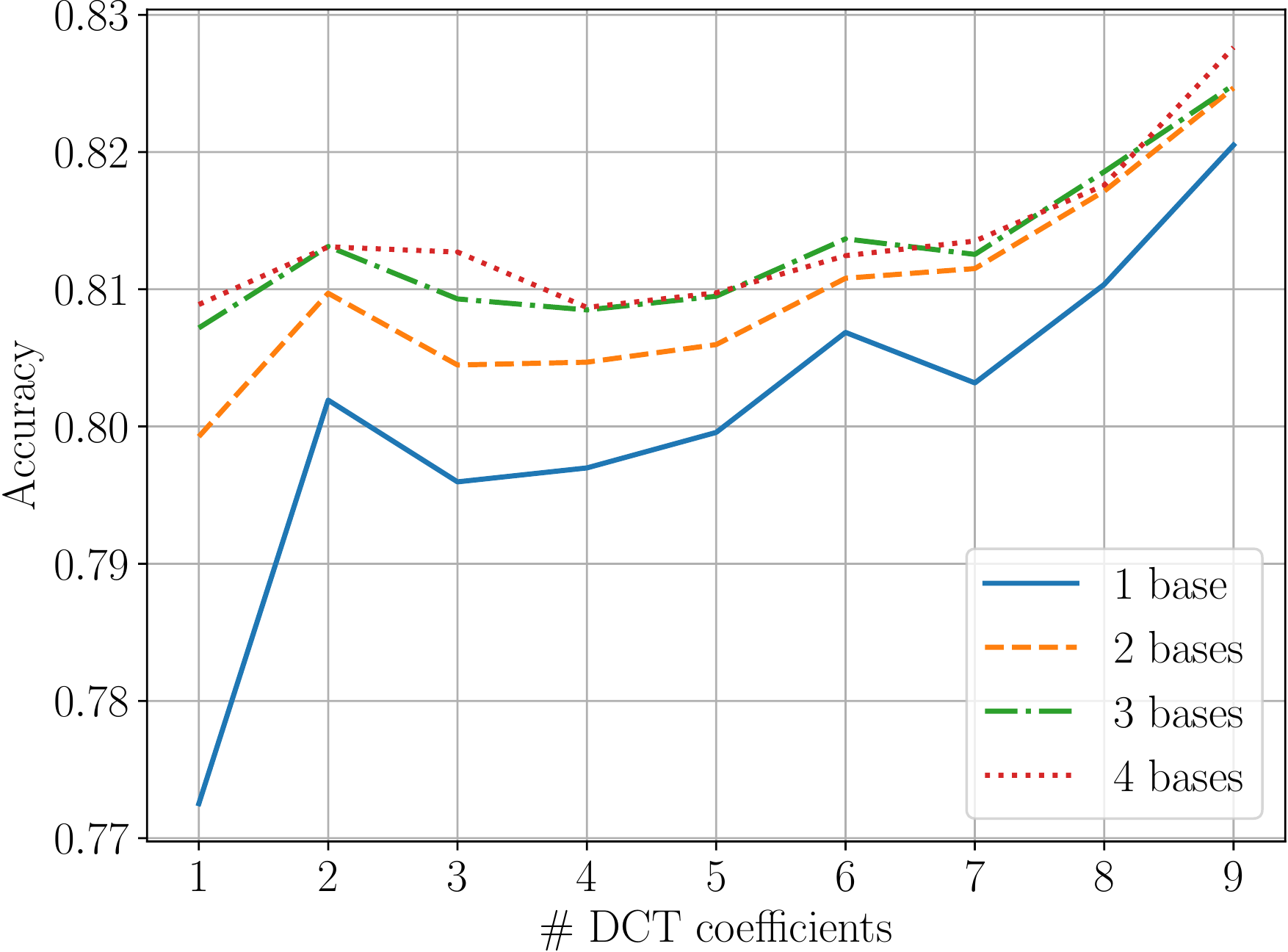}} \hfill
	\subfloat[all $\Delta$ values]{\includegraphics[width = 0.48\columnwidth]{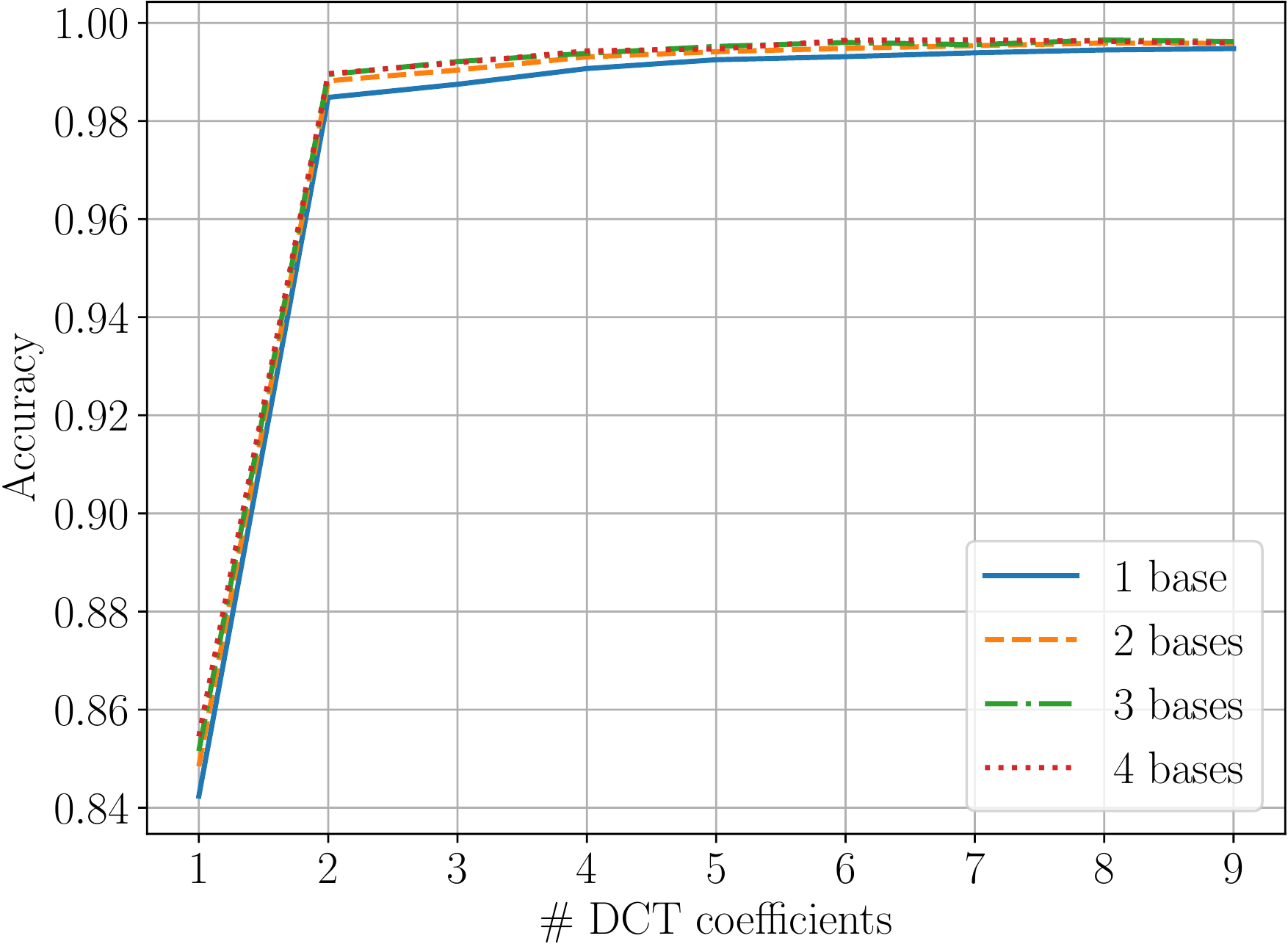}} \\
		
	\subfloat[single $b$ value]{\includegraphics[width = 0.48\columnwidth]{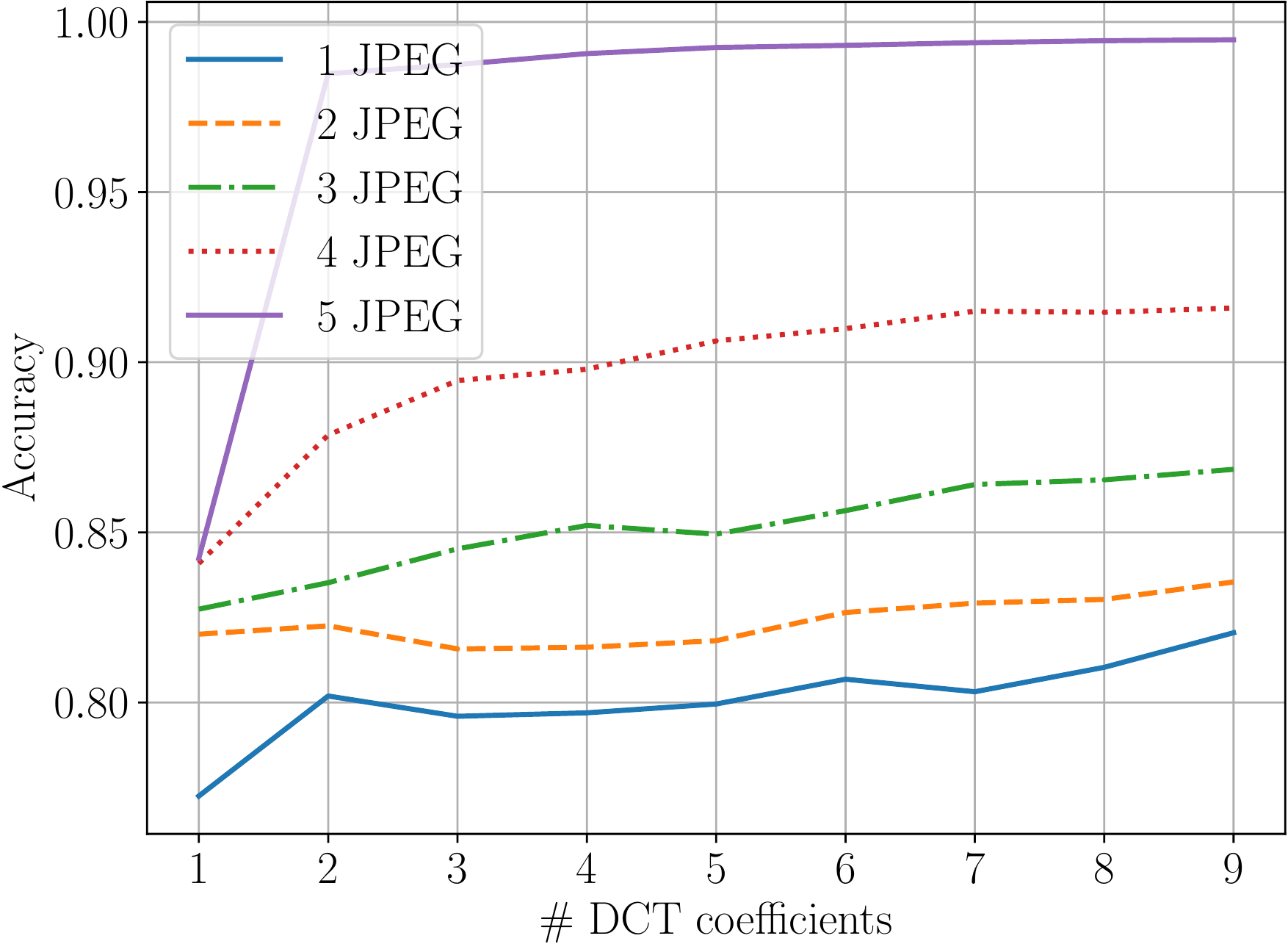}} \hfill
	\subfloat[all $b$ value]{\includegraphics[width = 0.48\columnwidth]{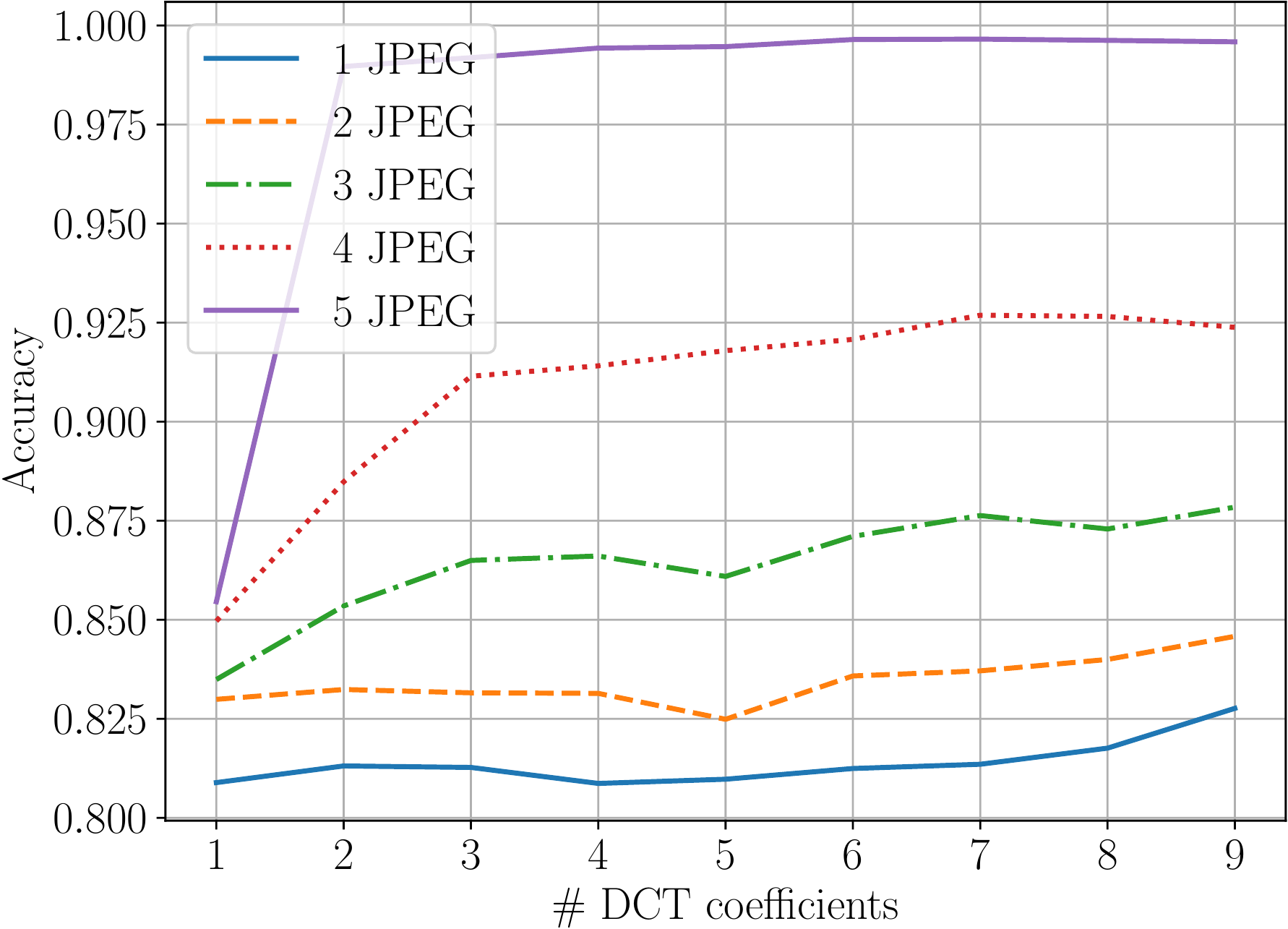}} \\
		
	\subfloat[single $n$ value]{\includegraphics[width = 0.48\columnwidth]{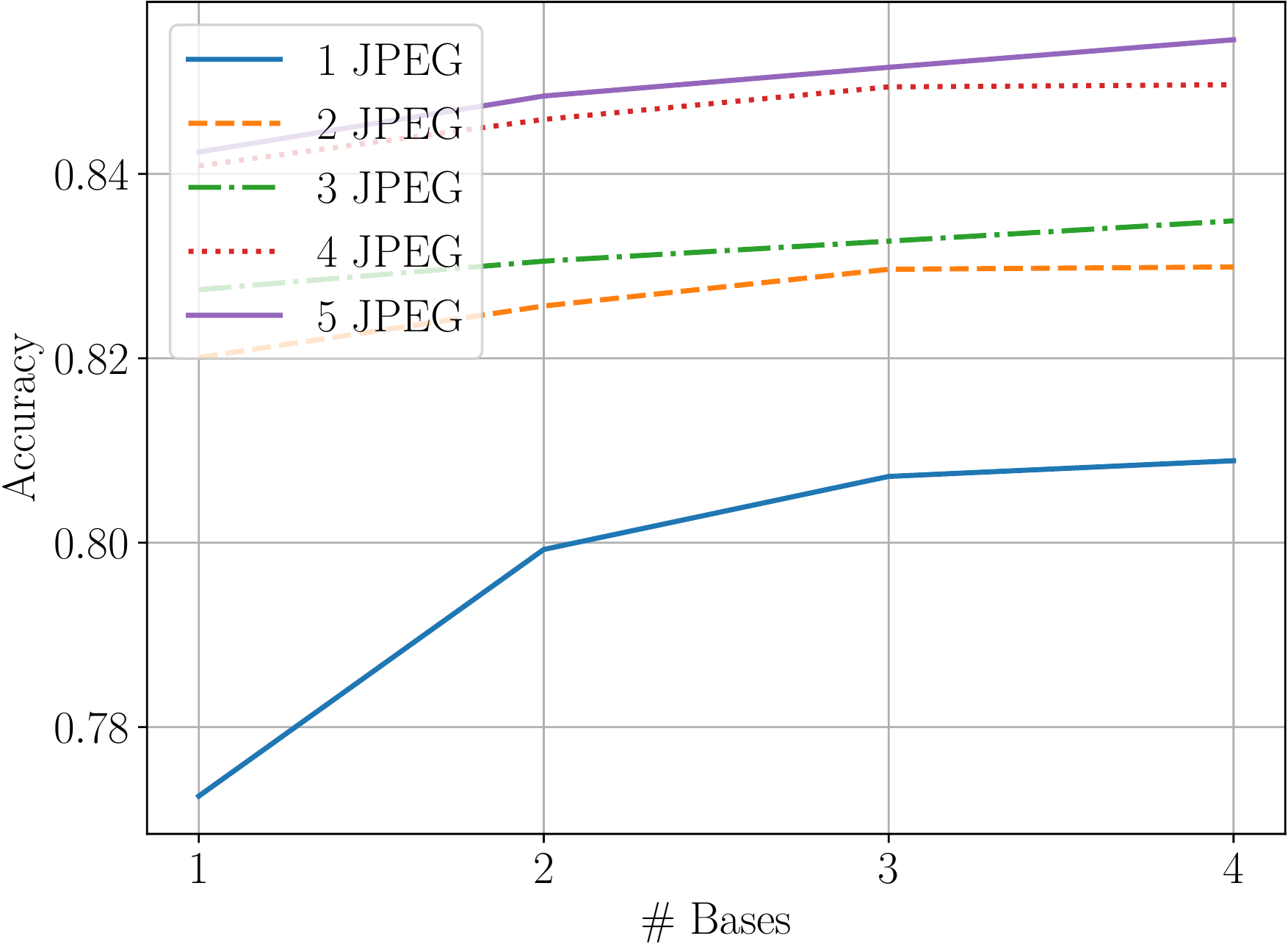}} \hfill
	\subfloat[all $n$ value]{\includegraphics[width = 0.48\columnwidth]{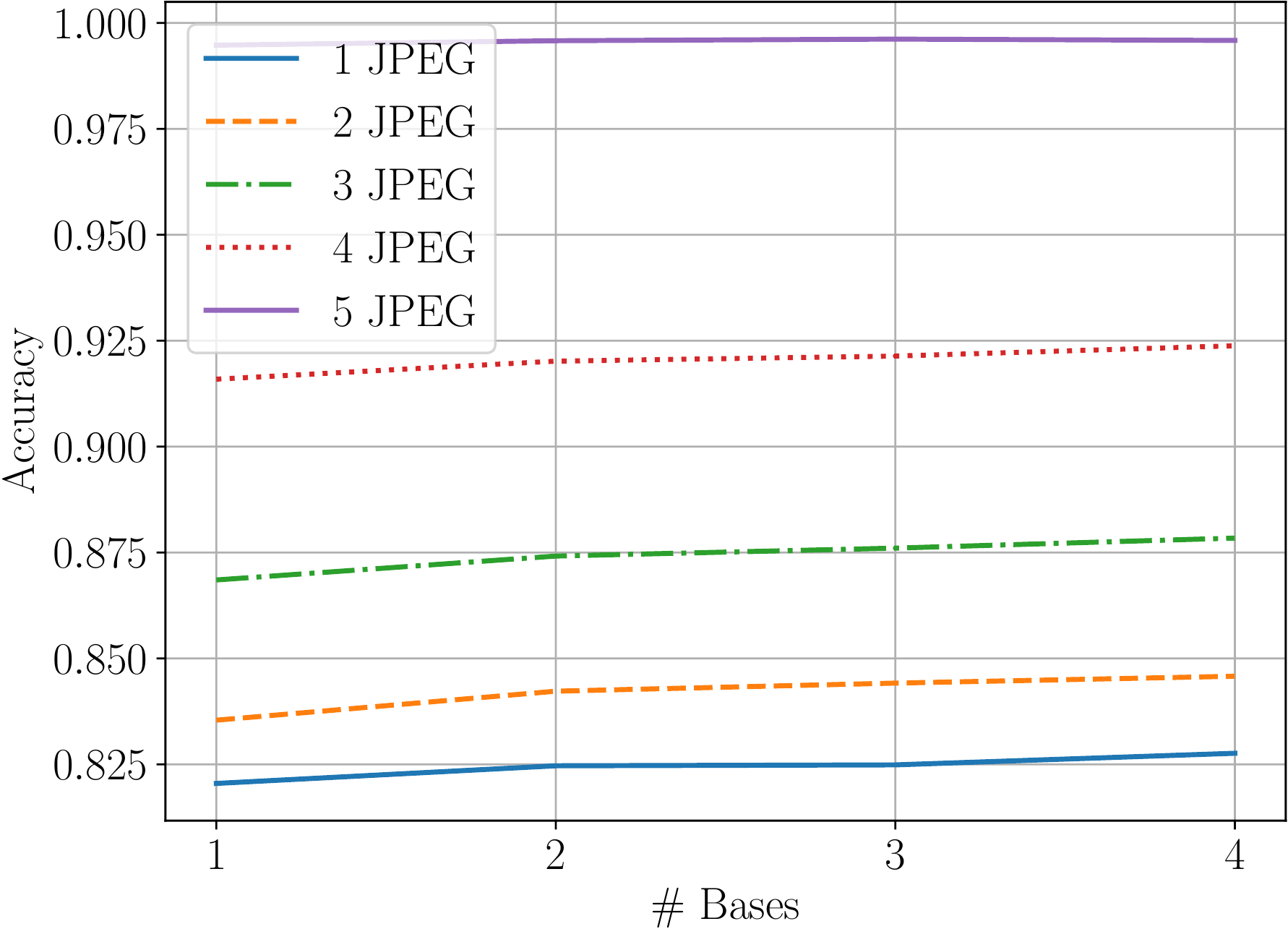}} 
	
	\caption{Accuracy varying different parameters: (a) fix $1$ JPEG compressions; (b) fix $5$ JPEG compressions; (c) fix $1$ base; (d) fix $4$ bases; (e) fix $1$ DCT coefficient; (f) fix $9$ DCT coefficients. When we fix a single parameter, we average all results obtained by fixing that parameter to each possible value it can span.}
	\label{fig:acc_fix_param}
\end{figure}


In order to gain a better insight on the effect of using different bases, \gls{dct} coefficients and quantization steps, we performed an analysis by keeping some parameters fixed, and just changing the others.
Fig.~\ref{fig:acc_fix_param}(a) and (b) show the results with a single fixed quantization step value, and considering all the values, respectively.
In both scenarios it is possible to notice that the greatest improvement is obtained when more than a single \gls{dct} coefficient is used.
Moreover, the more the coefficients, the better the results.
Fig.~\ref{fig:acc_fix_param}(c) and (d) report the results with features from a single fixed \gls{fd} base and from all the considered bases, respectively.
It seems that using more than one base only marginally improve the results.
As a matter of fact, both figures are very similar.
Finally, Fig.~\ref{fig:acc_fix_param}(e) and (f) display the accuracy values obtained from the features of a single \gls{dct} frequency and of the whole set of frequencies, respectively. .
From these figures it is possible to note that the more the considered quantization steps, the higher the accuracy for any other parameters set.
This is not particularly surprising, as Benford's law is naturally linked to the used JPEG quantization.

\vspace{.5em}\noindent\textbf{Comparison against baseline.}
In order to compare against the selected baselines solution, we finetuned Xception network and trained a linear \gls{svm} on the steganalysis features for each dataset according to the same procedure used for our Random Forest as suggested in \cite{Marra2018}. 
Table~\ref{tab:res_scen1} reports the breakdown of test accuracy scores for all datasets.
The highest average accuracy among the considered methods is obtained by the proposed method, and it is higher than $0.99$.
It is also interesting to notice that the proposed solution is considerably better than the baseline \gls{cnn} on \texttt{winter2summer}, \texttt{sats} and \texttt{lsun\_bedroom}, which seem to be particularly though for the latter.
These results highlight that, in order to properly train a very deep network like Xception, a much larger dataset probably is needed.
However, this might be difficult to obtain in a reduced amount of time in a forensic scenario.
On the contrary, the proposed feature vector is very compact, thus Random Forest does not suffer from a smaller training set.
The baseline handcrafted method performs reasonably well, but the obtained accuracy is lower than that of the proposed method of almost 9\% on average.

\begin{table}[t]
	\centering
	\caption{Accuracy results compared to the baseline solutions for each dataset. Average accuracy (avg) is also reported. Best result per dataset in bold.}
	\label{tab:res_scen1}
	\begin{tabular}{cccc}
		\toprule
		\textbf{Dataset}    & \textbf{Proposed} & \textbf{Xception} & \textbf{Steganalysis}\\ \midrule
		orange2apple        & $\bm{98.13} $             & $97.64          $                 & $88.80          $                 \\
		photo2ukiyoe        & $\bm{100.00}$             & $97.41          $                 & $86.78          $                 \\
		winter2summer       & $\bm{100.00}$             & $68.33          $                 & $77.96          $                 \\
		zebra2horse         & $\bm{99.69} $             & $89.58          $                 & $91.01          $                 \\
		photo2cezanne       & $\bm{99.97} $             & $95.91          $                 & $95.88          $                 \\
		photo2vangogh       & $\bm{100.00}$             & $93.75          $                 & $94.68          $                 \\
		photo2monet         & $\bm{99.84} $             & $94.08          $                 & $94.80          $                 \\
		facades             & $\bm{100.00}$             & $99.84          $                 & $73.93          $                 \\
		cityscapes          & $\bm{100.00}$             & $\bm{100.00}$                 & $\bm{100.00}$                 \\
		sats                & $\bm{99.69} $             & $73.00          $                 & $90.92          $                 \\
		lsun\_bedroom       & $\bm{100.00}$             & $76.22          $                 & $98.92          $                 \\
		lsun\_bridge        & $\bm{99.89} $             & $82.49          $                 & $95.90          $                 \\
		lsun\_churchoutdoor & $\bm{99.99} $             & $99.79          $                 & $98.81          $                 \\
		lsun\_kitchen       & $\bm{99.99} $             & $87.26          $                 & $99.49          $                 \\
		lsun\_tower         & $\bm{99.98} $             & $95.45          $                 & $98.87          $                 \\ \midrule
		avg                 & $\bm{99.83} $             & $89.64          $                 & $91.03          $                 \\ \bottomrule
	\end{tabular}
\end{table}

%

\vspace{.5em}\noindent\textbf{Resilience to JPEG compression.}
When images are shared online, JPEG compression is almost always applied in order to reduce network and storage requirements.
Therefore, we measured the performance of the proposed method whenever a further JPEG compression is applied with different coding parameter configurations.

In a first scenario, \gls{gan}-generated and real images have been randomly JPEG compressed considering quality factors distributed in $\left\{85, \ldots, 100\right\}$.
The originally-trained detector (on non-compressed images) was then tested on this newly compressed dataset
In this situation, the proposed solution approaches a random guess accuracy.
However, this situation is not completely unexpected.
As a matter of fact, Benford's law is strictly tailored to JPEG compression.
Therefore, scrambling with JPEG coefficients statistics through recompression has a high impact on Benford's features.

We therefore considered a second scenario, which is more realistic as shown in \cite{Marra2018}.
If we know that images might be JPEG recompressed, we can also train our system on JPEG compressed images.
We therefore re-trained our method and the baseline on compressed images, and tested them on compressed images.
In this situation, results improve as expected.
As a matter of fact, the proposed solution accuracy decreases, but still remains higher than $0.80$.
In particular, results depend on the specific datasets and \gls{gan} architecture.
Indeed, all results related to ProGAN (i.e., last five datasets) show an almost optimal accuracy always higher than $0.99$.
Conversely, on Cycle-Gan images, only a couple of datasets exhibit accuracy grater than $0.70$.
In this situation, if computational is feasible for the adopted architecture, the baseline network might be preferable.

We then tested a third scenario, assuming that the analyst knows which is the quality factor adopted by the final JPEG compression stage (since it can be read from the bit stream).
It is possible to train a different Random Forest classifier or Xception network for each quality factor.
We therefore generated three versions of the dataset by recompressing it with quality factor $100$, $95$, and $90$, respectively.
For each quality factor, we trained the proposed method and Xception baseline using the aforementioned leave-one-group-out strategy.
We did not consider steganalysis features anymore, as in \cite{Marra2018} the authors already showed that they greatly suffer JPEG compression.
Results are reported in Table~\ref{tab:res_jpeg}.
It is possible to notice that for high quality factors, the proposed Benford-based method outperforms the baseline.
Xception network shows better results starting from quality factor $90$.

In the final testing scenario, we assume that the analyst wants to train a different classifier for each JPEG quality factor, and for each kind of image content.
As an example, if the analyst is interested in detecting fake oranges with a given quality factors, he/she might train only on the orange2apple dataset, rather than the others.
In this situation (i.e., known quality factor and kind of GAN training dataset), both the proposed method and the Xception baseline achieve an almost perfect result for each quality factor (i.e., $100$, $95$ and $90$).

\begin{table}[t]
	\centering
	\caption{Accuracy obtained using different JPEG quality factors.}
	\label{tab:res_jpeg}
	\begin{tabular}{@{}cccc@{}}
		\toprule
		\textbf{QF}          & \textbf{Dataset}   & \textbf{Proposed} & \textbf{Xception} \\ \midrule
		\multirow{4}{*}{$100$} & orange2apple       & $\bm{94.50} $                      & $92.56     $                   \\
		                     & photo2ukiyoe       & $\bm{100.00}$                      & $98.50     $                   \\
		                     & cityscapes         & $\bm{100.00}$                      & $100.00    $                   \\
		                     & lsun\_tower        & $\bm{100.00}$                      & $94.64     $                   \\ \midrule
		\multirow{4}{*}{$95$}  & orange2apple       & $82.01      $                      & $\bm{90.66}$                        \\
		                     & photo2ukiyoe       & $97.00      $                      & $\bm{98.42}$                        \\
		                     & cityscapes         & $\bm{99.99} $                      & $99.32     $                   \\
		                     & lsun\_tower        & $\bm{99.80} $                      & $99.48     $                   \\ \midrule
		\multirow{4}{*}{$90$}  & orange2apple       & $65.93      $                      & $\bm{85.61}$                        \\
		                     & photo2ukiyoe       & $92.01      $                      & $\bm{98.17}$                        \\
		                     & cityscapes         & $\bm{100.00}$                      & $99.66     $                   \\
		                     & lsun\_tower        & $99.60      $                      & $\bm{98.86}$                        \\ \bottomrule
	\end{tabular}%
\end{table}

\subsection{Analysis on faces.}
All results shown so far are obtained not considering GANs generating face images.
This is due to two main reasons.
First, GANs that were trained to generate face images produce particularly realistic results lately.
This make face images harder to detect as GAN-generated compared to other kind of imagery.
Indeed, shadows and lightning very often respect physics law, thus making Benford's law almost verified \cite{acebo2015}.
Second, face-generating GANs are often trained on common pristine face datasets~\cite{celeb-a}, which makes the Leave-One-Group-Out testing strategy applied to now impracticable.

In the light of these considerations, we decided to create a specific dataset, composed by all the faces dataset in the original corpus, generated by ProGAN~\cite{karras2018} (\num{19870} images), StarGAN~\cite{Choi2018stargan} (\num{50000} images) and GlowGAN~\cite{Kingma2018Glow} (\num{49900} images), plus some additional images generated by the more recently proposed StyleGAN2~\cite{Karras2019stylegan2} (\num{2000} images).
As pristine faces, we always consider images from the Celeb-A dataset~\cite{celeb-a}.

\begin{figure}[t]
	\centering
	\subfloat{\includegraphics[width=0.15\linewidth]{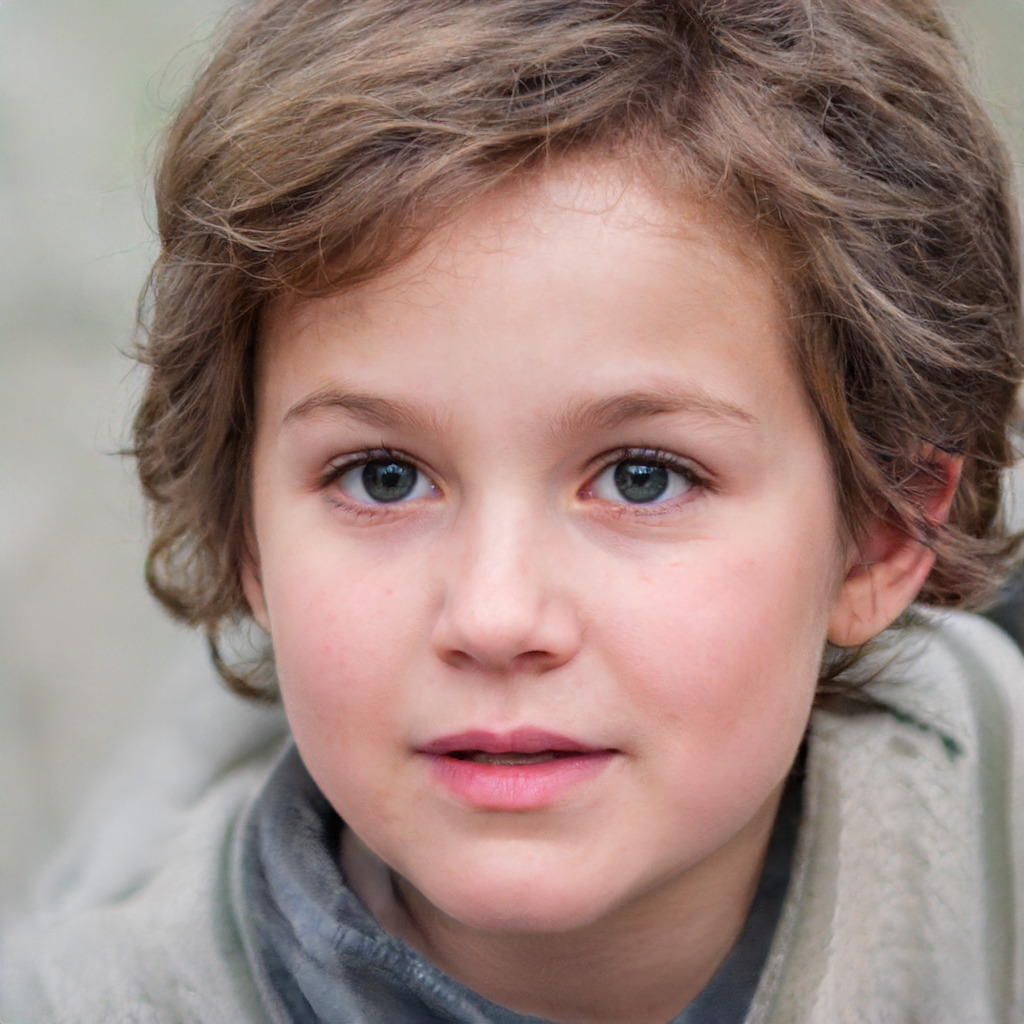}} \hfil
	\subfloat{\includegraphics[width=0.15\linewidth]{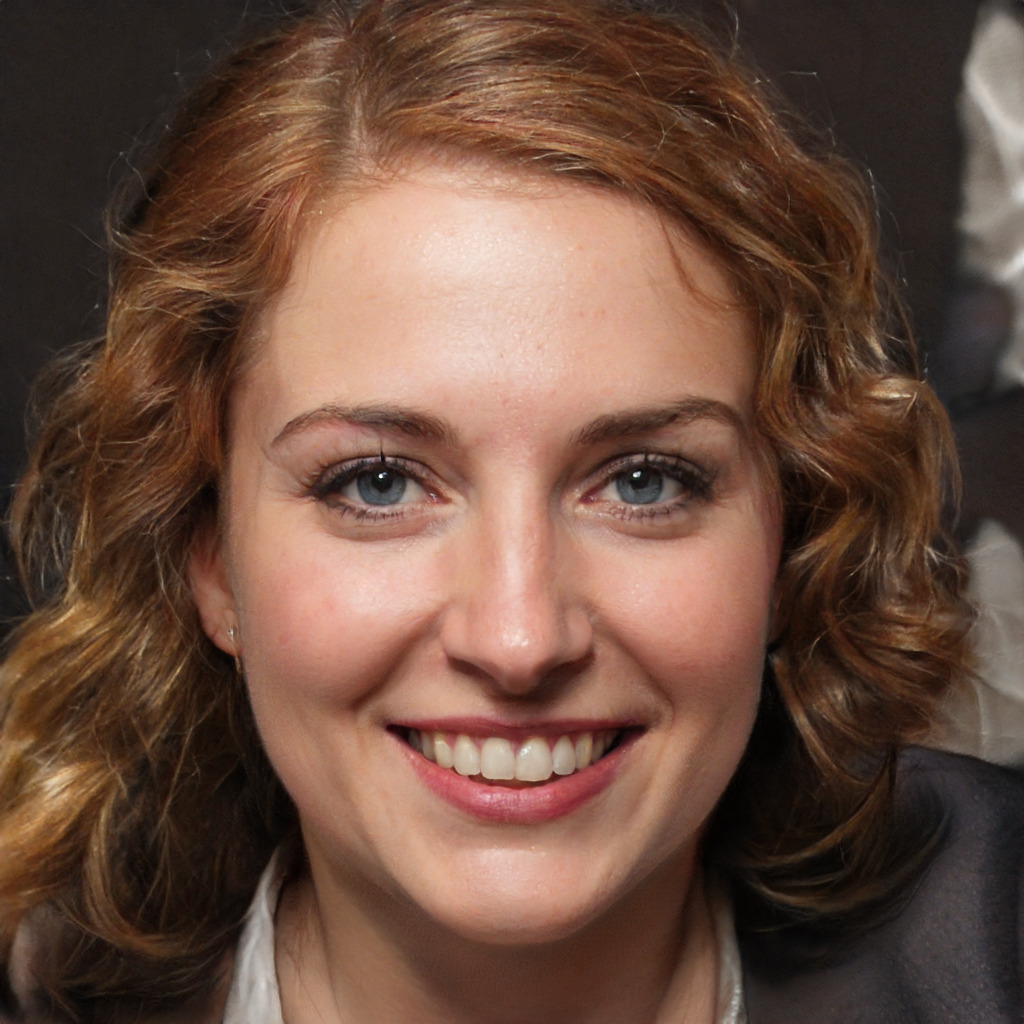}} \hfil
	\subfloat{\includegraphics[width=0.15\linewidth]{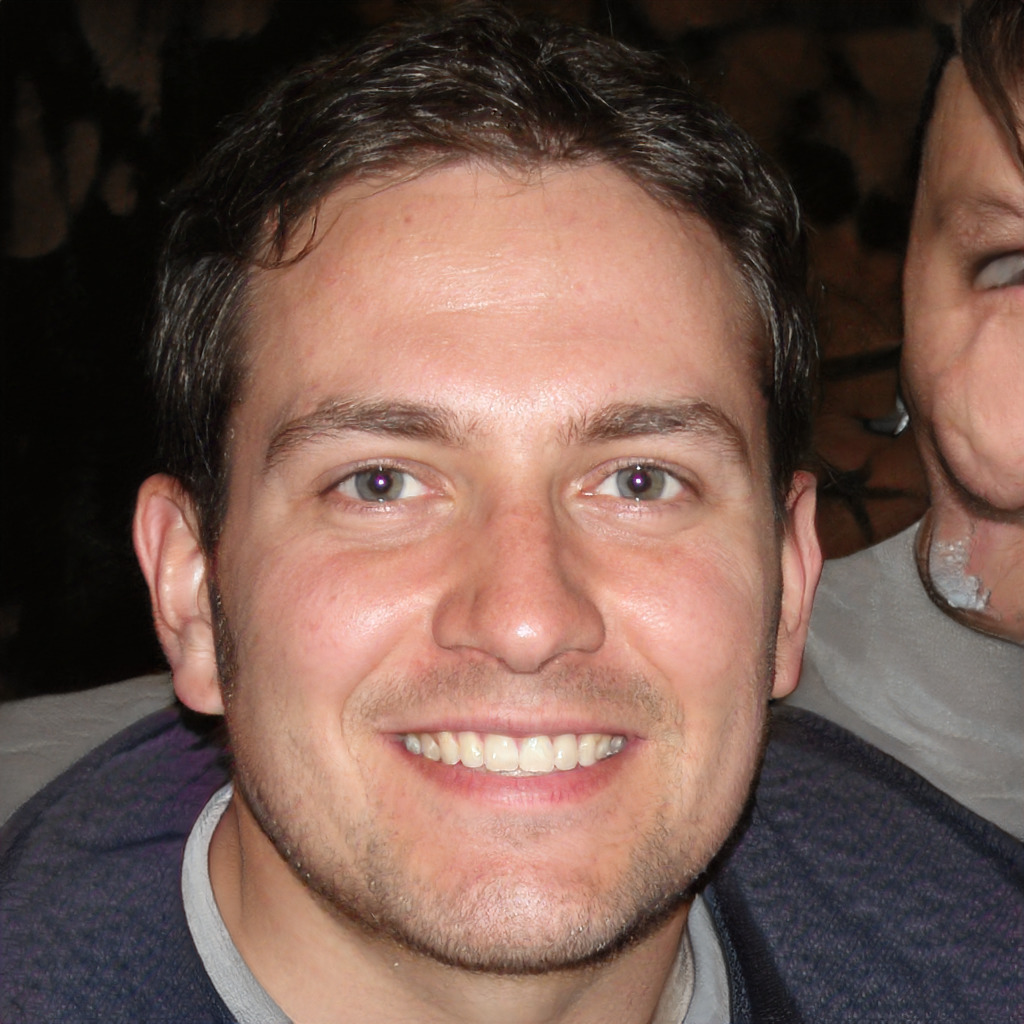}} \hfil
	\subfloat{\includegraphics[width=0.15\linewidth]{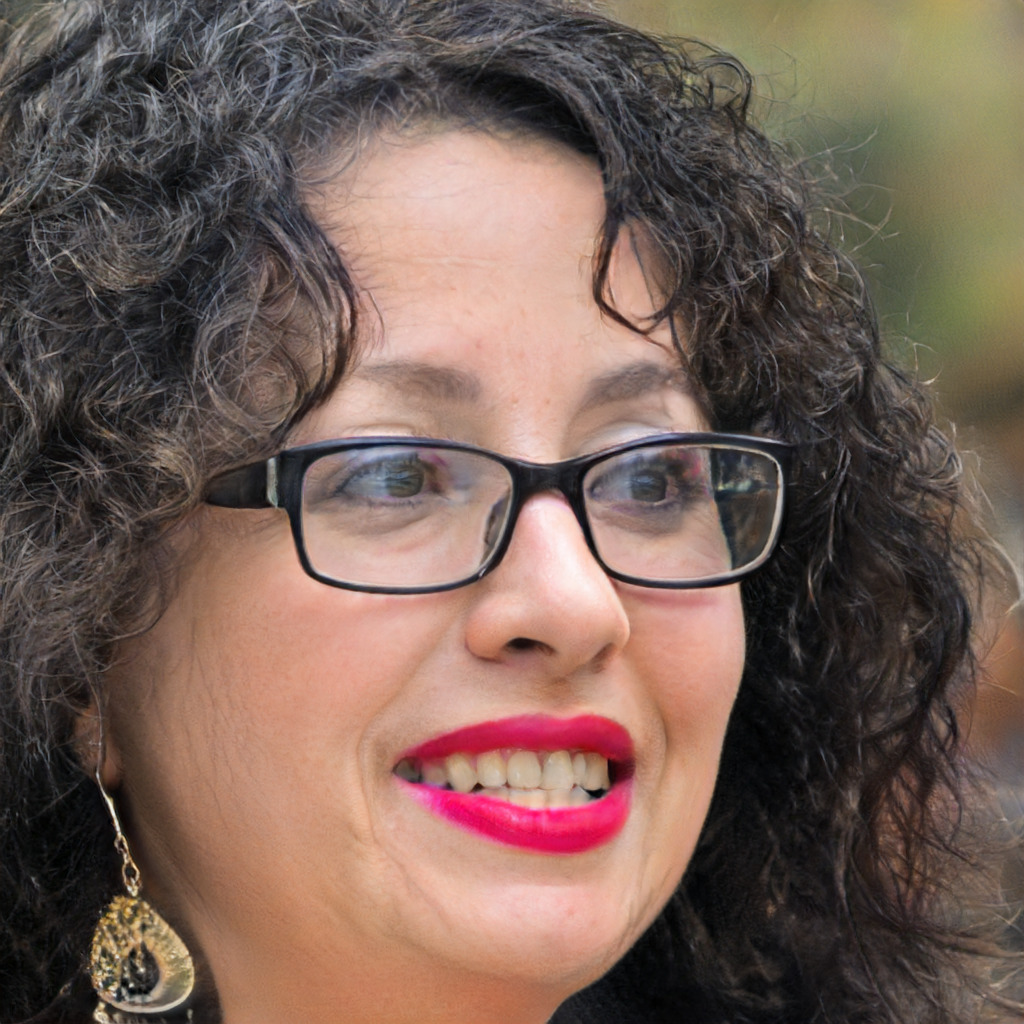}} \hfil
	\subfloat{\includegraphics[width=0.15\linewidth]{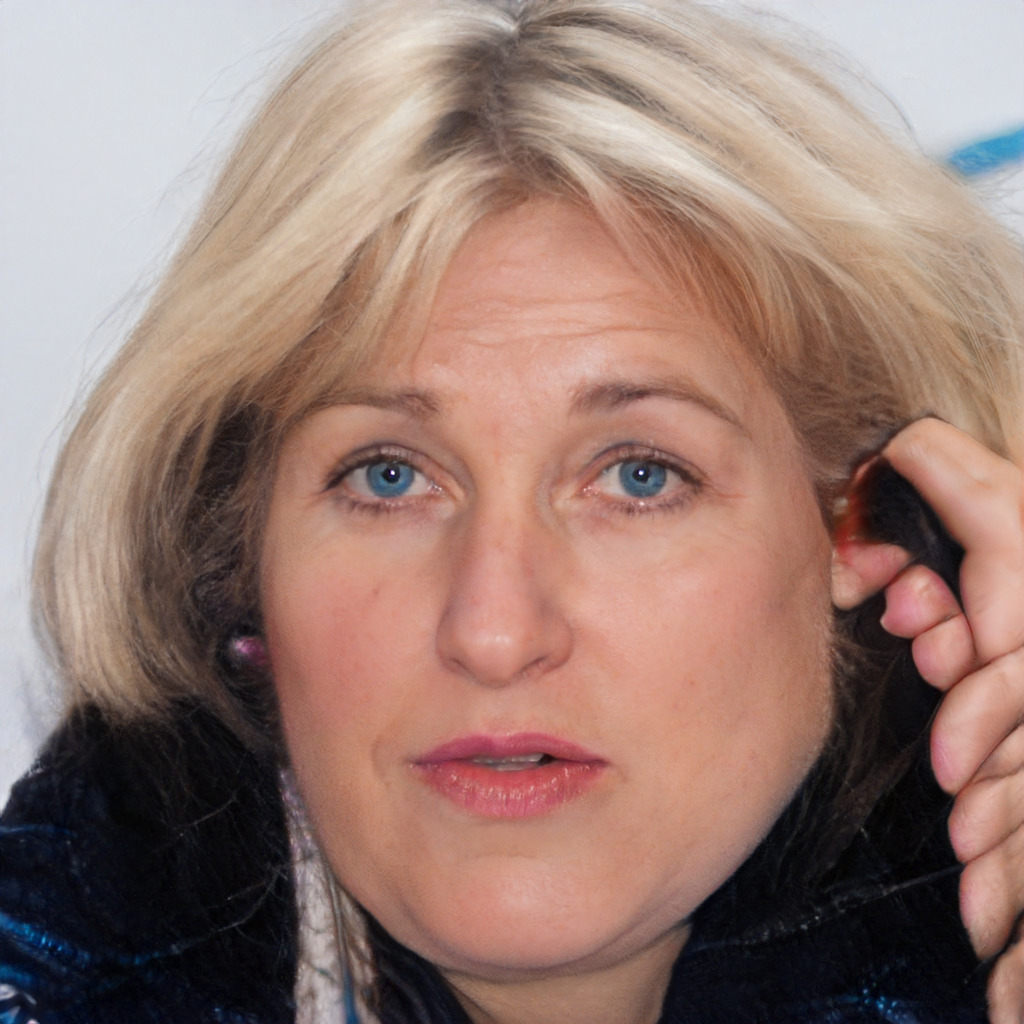}} \hfil
	\subfloat{\includegraphics[width=0.15\linewidth]{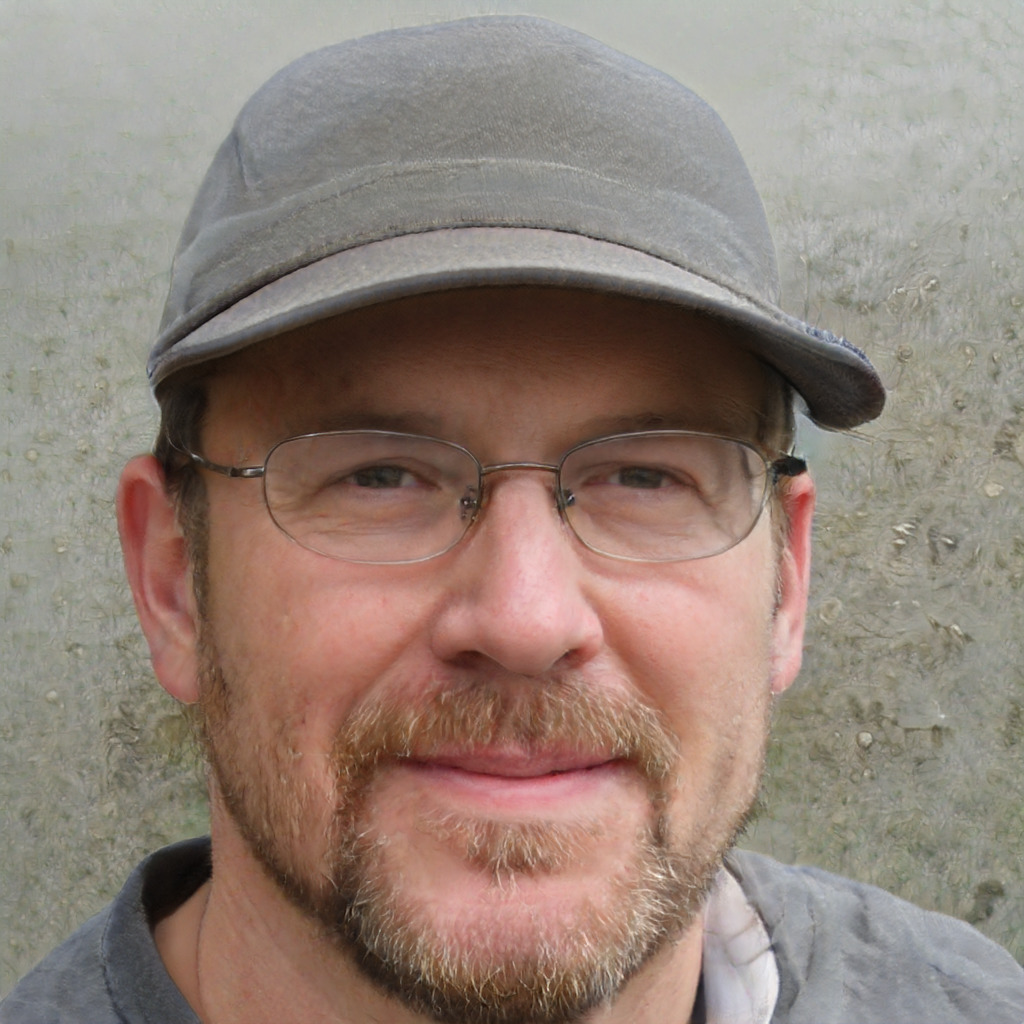}} \hfil
	\caption{Examples of faces generated with StyleGan2~\cite{Karras2019stylegan2}.}
	\label{fig:example_faces}
\end{figure}

ProGAN is trained to generate realistic faces similar to those from Celeb-A dataset~\cite{celeb-a}.
StarGAN and GlowGAN are trained to obtain faces with different characteristics (e.g., hair colors, smiles, etc.).
Finally, StyleGAN2 produces images at different qualities depending on its configuration parameter $\psi=0.5$ or $\psi=1$ as suggested by the authors~\cite{Karras2019stylegan2}, starting from the Flickr-Faces-HQ Dataset~\cite{karras2018}.
Some random images from those generated by StyleGAN2 are shown in Fig.~\ref{fig:example_faces}.
For each dataset, we train a Random Forest classifier considering 70\% of the images as training set and 30\% as test.

Table~\ref{tab:res_faces} shows the achieved results on each test set. It is possible to notice that StarGAN and GlowGAN seems to be easier to detect.
On the contrary, ProGAN and StyleGAN2 looks more challenging.
These promising preliminary results motivate some future work with more extended face image datasets, also comparing against other baselines and in presence of editing operations.
%
%
\begin{table}[t]
	\centering
	\caption{Accuracy results for each face test dataset. Average accuracy (avg) is also reported. Accuracy higher than $85\%$ are reported in bold.}
	\label{tab:res_faces}
	\begin{tabular}{cc}
		\toprule
		\textbf{Dataset}	     & \textbf{Proposed}	 \\ \midrule
		progan\_celeba           & $79.75$                       \\
		stargan\_black\_hair     & $97.26$                       \\
		stargan\_blond\_hair     & $96.56$                \\
		stargan\_brown\_hair     & $96.76$                \\
		stargan\_male            & $96.24$               \\
		stargan\_smiling         & $96.06$              \\
		glow\_black\_hair        & $86.56$                     \\
		glow\_blond\_hair        & $88.26$                     \\
		glow\_brown\_hair        & $86.18$                     \\
		glow\_male               & $87.11$              \\
		glow\_smiling            & $83.04$               \\
		stylegan2-0.5            & $77.18$                      \\
		stylegan2-1              & $72.63$                      \\ \midrule
		avg                      & $87.96$                       \\ \bottomrule
	\end{tabular}
\end{table}
\section{Conclusions}\label{sec:conclusions}
In this paper we proposed a study on the use of the well-known Benford's law for the task of \gls{gan}-generated image detection.
We proposed a strategy to extract Benford-related features from an image relying on different divergence definitions.
We also showed how to combine these features in order to better exploit different bases as well as \gls{dct} frequencies.
Using these features, we performed a series of experiment based on a simple Random Forest classifier in order to study the amount of information captured by the features, rather than focusing on specializing a complex classifier.

Results show that \gls{gan}-generated images often fail in respecting Benford's law, thus can be discriminated from natural pictures.
However, some kind of \gls{cnn} architectures seem to produce images that are harder to detect than others.
This motivates future studies on this topic.
Moreover, this suggests to possibly embed Benford's law into \gls{gan} loss function in order to obtain even more realistic images.

\bibliographystyle{IEEEtran}
\balance
\bibliography{biblio}

\end{document}